\pdfoutput=1

\documentclass[11pt]{article}

\usepackage[]{ACL2023}

\usepackage{times}
\usepackage{latexsym}

\usepackage[T1]{fontenc}

\usepackage[utf8]{inputenc}

\usepackage{microtype}

\usepackage{inconsolata}

\usepackage{mathtools}
\usepackage{amsmath}
\usepackage{booktabs}
\usepackage{makecell}
\usepackage{graphicx}
\usepackage{svg}
\usepackage{float}

\usepackage{listings}

\DeclareMathOperator*{\argmax}{argmax}

\graphicspath{./images/}

\title{\textsc{Syndicom}: Improving Conversational Commonsense \\with Error-Injection and Natural Language Feedback}

\author{Christopher Richardson, Anirudh Sundar, Larry Heck \\
  School of Electrical and Computer Engineering \\
  Georgia Institute of Technology \\
  Atlanta, GA, 30308, USA \\
  \texttt{\{crichardson8,asundar34,larryheck\}@gatech.edu}}

\begin{document}
\maketitle
\begin{abstract}
Commonsense reasoning is a critical aspect of human communication. Despite recent advances in conversational AI driven by large language models, commonsense reasoning remains a challenging task. In this work, we introduce \textsc{Syndicom} - a method for improving commonsense in dialogue response generation. \textsc{Syndicom} consists of two components. The first component is a dataset composed of commonsense dialogues created from a knowledge graph and synthesized into natural language. This dataset includes both valid and invalid responses to dialogue contexts, along with natural language feedback (NLF) for the invalid responses. The second contribution is a two-step procedure: training a model to predict natural language feedback (NLF) for invalid responses, and then training a response generation model conditioned on the predicted NLF, the invalid response, and the dialogue.

\textsc{Syndicom} is scalable and does not require reinforcement learning. Empirical results on three tasks are evaluated using a broad range of metrics. \textsc{Syndicom} achieves a relative improvement of 53\% over ChatGPT on ROUGE-1, and human evaluators prefer \textsc{Syndicom} over ChatGPT 57\% of the time. We will publicly release the code and the full dataset.

\end{abstract}
\section{Introduction}

\begin{figure*}
\centering
\includegraphics[width=0.83\textwidth]{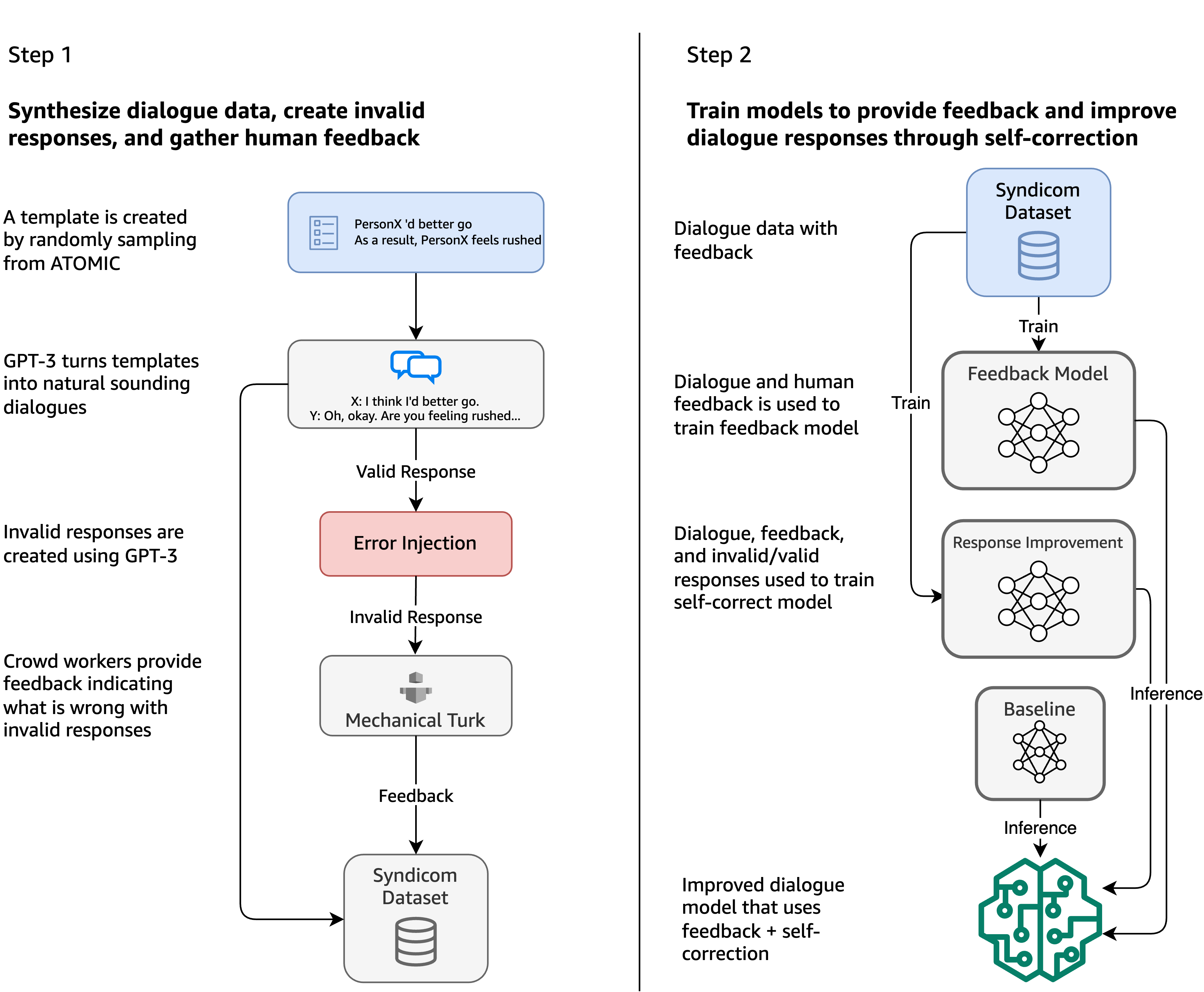}
\caption{\textsc{Syndicom} Process. Left: dataset generation, Right: Improving commonsense in dialogue response generation.}
\label{fig:syndicom}
\end{figure*}

Conversational AI has witnessed rapid advancements in recent years, largely due to the success of large language models (LLMs) such as GPT-3 \cite{brown2020language}. These advancements have been driven by the notable achievements of models like ChatGPT, which is built upon InstructGPT \cite{ouyang2022training}. InstructGPT was trained on an extensive dataset of instructions for various language tasks and was further enhanced using human feedback and reinforcement learning (RL). Consequently, research in conversational AI has shifted towards leveraging large models trained on extensive datasets, supplemented by human feedback.

While these models have consistently demonstrated significant improvements in reasoning and problem-solving capabilities, they still exhibit flaws and issues. In many critical applications of LLMs, the tolerance for errors in dialogue responses is exceedingly low. Addressing these problems remains challenging, primarily due to the scarcity of data and the high cost associated with human feedback. Recent research has started exploring alternative techniques beyond human feedback and RL, such as natural language feedback (NLF) and self-correction \cite{saunders2022self, scheurer2022training, welleck2022generating, bai2022constitutional}.

Furthermore, even with the progress made, large models often generate hallucinations, underscoring the ongoing importance of knowledge grounding. One of the most demanding aspects of knowledge grounding is commonsense knowledge. Recent advancements in incorporating commonsense into LLMs have utilized resources such as ConceptNet \cite{speer2017conceptnet} or \textsc{Atomic} \cite{sap2019atomic}.

This paper presents a method for improving commonsense dialogue responses by (1) replacing human feedback and RL with natural language responses and (2) leveraging recent knowledge graph techniques to ground responses in commonsense knowledge derived from \textsc{Atomic}. To address the scarcity of data and the high cost of human feedback, the natural language feedback is elicited in a manner that specifically targets the chosen error types determined by the designer. This approach significantly enhances the speed and quality of model learning and refinement.

The contributions of this paper are as follows:
\begin{itemize}
\item Development of a scalable method for synthesizing knowledge-grounded data with error injection and feedback.
\item Release of a dataset rich in dialogues featuring commonsense inferences, annotated with commonsense errors, and accompanied by human-written feedback, which we refer to as \textsc{Syndicom}.
\item Description of a method for training both a feedback generation model and a response improvement model using natural language feedback (NLF), and demonstration of the superiority of this information-rich approach over state-of-the-art RL methods using \textsc{Syndicom}.
\end{itemize}

\section{Recent Work}

The field of conversational AI has experienced a surge of interest in commonsense reasoning in recent years, with a significant focus on curating datasets \cite{richardson2023commonsense}. ConceptNet \cite{speer2017conceptnet} and ATOMIC \cite{sap2019atomic} have emerged as widely used resources for dataset curation, establishing a de facto standard. Several datasets serve as sources for the dialogues, including DailyDialogue \cite{li2017dailydialog}, MuTual \cite{cui2020mutual}, DREAM \cite{sun2019dream}, and the Ubuntu Dialogue Corpus \cite{lowe2015ubuntu}.

Our research lies at the intersection of two critical areas in conversational AI: the synthesis of commonsense datasets and the training of models using natural language feedback. These areas have recently garnered significant research attention due to their potential to enhance the ability of conversational agents to understand and respond to complex human interactions with greater accuracy and consistency. By leveraging the synergies between these domains, our work aims to address the existing limitations in conversational agents and pave the way for more robust and effective conversational systems.

\subsection{Commonsense Dataset Curation}

In recent years, various datasets have been curated specifically for commonsense reasoning. Ghosal et al. (2021) introduced CIDER, a dialogue dataset annotated with commonsense inferences, which was later expanded with the more open-ended CICERO \cite{ghosal-etal-2022-cicero}. Some researchers have focused on specific types of commonsense, such as temporal commonsense \cite{qin2021timedial} and ethical commonsense \cite{ziems2022moral, kim2022prosocialdialog, sun2022moraldial}. Others have concentrated on grounding dialogues in knowledge graphs \cite{zhou2021commonsense, moon2019opendialkg}.

These approaches rely on existing dialogue datasets and often employ filtering strategies to reduce dataset size. However, this reliance on existing datasets can limit the generalizability of methods to future problems. One potential solution to the scarcity of large-scale annotated commonsense knowledge datasets is the synthesis approach. Recently, Kim et al. (2022) proposed \textsc{Soda}, a method for procedurally generating social dialogues based on a commonsense knowledge graph. They utilized ATOMIC \cite{sap2019atomic}, which consists of atomic facts in natural language form, to generate synthetic dialogues rich in commonsense inferences. Their entirely procedural and highly scalable approach generates dialogue data suitable for training models that reason over commonsense knowledge. Building upon this work, we present \textsc{Syndicom}, a synthesis procedure and dataset that expands on the ideas of \textsc{Soda} and incorporates novel features crucial for our dialogue modeling approach. More details about \textsc{Syndicom} are provided in Section \ref{sec:syndicom}.

\subsection{Feedback and Response Improvement}

The use of feedback to improve language models has recently garnered increased interest, with most efforts focused on the application of reinforcement learning \cite{stiennon2020learning, zhou2021narle, bai2022training, bai2022constitutional}. Reinforcement learning with human feedback (RLHF) is particularly notable as it serves as the foundation for InstructGPT \cite{ouyang2022training}, which paved the way for ChatGPT. RLHF offers a flexible approach to improving LLMs; however, it faces challenges in terms of stability and efficiency inherent to RL. Moreover, the low dimensionality of the reward signal in RL (typically a scalar) severely limits the learning rate.

A more information-rich approach than RL is the use of natural language feedback (NLF). NLF has been explored in several recent works. Scheurer et al. (2022) investigated the use of human-written NLF to train a dialogue response refinement model. Saunders et al. (2022) demonstrated that LLMs themselves can generate this feedback. Welleck et al. (2022) developed a method to improve sequence generation of LLMs by first generating a baseline using an imperfect base generator and then correcting the output using a second correction model. The correction model incorporates feedback as part of its input. However, the authors only demonstrated the use of feedback provided by various tools and APIs tailored to the specific tasks they explored.
\section{The \textsc{Syndicom} Method} \label{sec:syndicom}
\begin{table*}
\tiny
    \centering
    \begin{tabular}{l|l|l}
    \toprule
    \textbf{Template} & \textbf{Synthesized Dialogue} & \textbf{Explanation} \\
    \midrule
    \thead[lt]{PersonX refuses PersonY\\$\hookrightarrow$ PersonX is seen as: disagreeable\\$\hookrightarrow$ As a result, PersonX feels: annoyed and irritated\\$\hookrightarrow$ Before that, PersonX needed: thinks about it} & \thead[lt]{\textbf{Context}\\I refuse to do what you ask.\\\textit{Why are you being so disagreeable?}\\I'm just annoyed and irritated.\\\textbf{Valid Response:}\\\color{teal}You should think about it before you say no.\\\textbf{Invalid Response:}\\\color{orange}You should think about it before you say yes.} & \thead[lt]{\textbf{Crowd Worker 1:}\\The other person already said no.\\\textbf{Crowd Worker 2:}\\The person did not say yes so this response was strange.} \\
    \midrule
    \thead[lt]{PersonX makes music\\$\hookrightarrow$ As a result, PersonX wants: to impress\\$\hookrightarrow$ PersonX is seen as: talented\\$\hookrightarrow$ As a result, PersonX will: gets asked to play something\\$\hookrightarrow$ Before that, PersonX needed: to carry their violin\\$\hookrightarrow$ PersonX wanted: they want to share their creativity} & \thead[lt]{\textbf{Context}\\I love making music.\\\textit{You're very talented!}\\Thanks. I just want to impress people with my playing.\\\textit{I'm sure you will. Can you play something for me?}\\Of course! I always carry my violin with me.\\\textbf{Valid Response:}\\\color{teal}That's great. I'm glad you want to share your creativity.\\\textbf{Invalid Response:}\\\color{orange}That's awful. I don't want to share my creativity.} & \thead[lt]{\textbf{Crowd Worker 1:}\\This contradicts with what was said about impressing people.\\\textbf{Crowd Worker 2:}\\They aren't being asked to play. They asked the other person to play} \\
    \bottomrule
    \end{tabular}
    \caption{Example dialogues from \textsc{Syndicom}. Each dialogue context includes both valid and invalid responses, as well as crowd worker-written explanations for the invalid response.}
    \label{tab:synthesis}
\end{table*}

Taking inspiration from recent NLF methods, this paper presents a new approach called \textsc{Syndicom}. This new approach combines the synthesis of commonsense dialogue data from a grounded knowledge graph (\textsc{ATOMIC}) with an NLF response improvement approach to improve dialogue responses. Figure \ref{tab:synthesis} illustrates the two phase process.

\subsection{\textsc{Syndicom} Dataset}
The \textsc{Syndicom} dataset is created in a four step process: (1) Auto-generate commonsense dialogue templates, (2) Translate templates into natural language dialogues, (3) Generate invalid responses with error injection, and (4) Collect human-written explanations for the invalid responses. Examples from the \textsc{Syndicom} dataset are shown in Table \ref{tab:synthesis}. The GPT model we used for the steps in this section was \lstinline{text-davinci-003}. Statistics for the dataset are shown in Table \ref{tab:stats}.

\subsubsection{Generating Templates}

Our approach generates commonsense-focused dialogue templates from a commonsense knowledge base. For this study, we utilize ATOMIC \cite{hwang2021comet}. ATOMIC consists of inferences in the form of $\text{Head} \xrightarrow{\text{relation}} \text{Tail}$. Each head and tail is a natural language description of a generic event, emotional state, action, description, etc. Dialogue templates are constructed by crawling through inferences rooted at each head of ATOMIC and chaining these inferences together to form multiple dialogue turns. The number of dialogue template turns is uniformly and randomly chosen between 3 and 8. 

\subsubsection{Converting to Natural Language}

Given the dialogue templates, the second step in creating \textsc{Syndicom} converts the templates to natural language conversations. We explored several methods, including crowdsourcing, but found LLMs to be the most consistent and effective. We used the GPT LLM (text-davinci-003) to generate the natural language dialogues from the templates. This was followed by in-context learning with 15 hand-written examples. The exact prompting used is shown in detail in Appendix \ref{sec:appendix}.

\subsubsection{Error Injection}
\label{sec:error_injection}

To elicit feedback on commonsense from crowd workers, the \textsc{Syndicom} process starts by corrupting the valid dialogue responses so that they violate commonsense reasoning. This provides crowd workers with an easy target for their feedback. To corrupt the dialogue responses, \textsc{Syndicom} takes advantage of the commonsense dialogue inference structure provided by \textsc{Atomic}. Given a commonsense knowledge base $\mathcal{K}$, a dialogue context $\mathcal{C}$, and response $r$ from \textsc{Syndicom}, the response is implied by commonsense from the context, or $\mathcal{C} \xrightarrow{\mathcal{K}} r$. The response $r$ is corrupted by replacing it with the semantic opposite, $\overline{r}$. We prompted GPT as shown in Appendix \ref{sec:appendix} to acquire these semantic opposites. The result is dialogues annotated with commonsense contradictions of the form $\{\mathcal{C}, r, \overline{r}\}$.  

\subsubsection{Natural Language Feedback Acquisition}

The dialogues with commonsense contradictions are presented to crowd workers on the Amazon's Mechanical Turk platform. Each dialogue is shown in the form of context and invalid responses, informing them that the dialogues were generated by an AI attempting to sound human. The crowd workers were given instructions to review AI-generated casual text message conversations and provide 1-2 sentences of natural language feedback on the dialogue, and the final turn in particular (the invalid response). They were asked to be as specific as possible in their feedback. The full instructions and web interface given to the crowd workers can be found in Appendix \ref{sec:appendix}.

To ensure the quality of the feedback, we used only masters-level crowd workers from English-speaking countries. This decision aimed to maximize the clarity and accuracy of the feedback provided. Each dialogue was evaluated by two crowd workers independently, allowing for a more comprehensive understanding of the AI's mistakes and ensuring a diverse range of feedback.

With the addition of the feedback $f$, this completes the dataset synthesis part, resulting in annotated dialogues of the form $\{\mathcal{C}, r, f, \overline{r}\}$. 

\subsection{\textsc{Syndicom} Dialogue Improvement} 
\label{sec:dia_improvement}
This section details the process of using natural language feedback to correct latent errors in the baseline conversational response. To begin, the  dialogue response improvement problem is defined as follows: given a dialogue context $\mathcal{D}$ and a response $r_b$, generated by some dialogue system or model, produce an improved response $r^*$.
\begin{equation}
    r^* = \argmax_{r} p(r \vert \mathcal{D},r_b)
\end{equation}

Dialogue response generation and improvement has recently received considerable attention \cite{shah2016interactive,  nayak2017plan,liu2017end, liu-etal-2018-dialogue, weston2018retrieve}. This problem is especially relevant today with large language models (LLMs). While LLMs have recently reached a high degree of fluency in dialogue, in some domains they can be factually inaccurate. While these cases are relatively infrequent, the tolerance for factual errors for a number of important applications is very low. In addition, these errors are difficult to predict and/or automatically detect.  This leads to a problem of data sparsity that is difficult to overcome for response improvement methods that rely on training models. 

A method to partially mitigate the sparsity of dialogue response errors is to {\em artificially create} \textit{invalid responses} $\overline{r}$ via error injection (as described in Section \ref{sec:error_injection}). This method will be called \mbox{\textsc{Syndicom-Direct}}. Given the invalid response $\overline{r}$ and the dialogue history $\mathcal{D}$, a model is trained to learn the optimal response $r^{*}$
\begin{equation}
    r^* = \argmax_{r} p(r \vert \mathcal{D},\overline{r}).
    \label{eq:direct}
\end{equation}

A second approach called \textsc{Syndicom-NLHF} includes natural language human feedback (NLHF) to explain the rationale for why the response $\overline{r}$ is invalid and then conditions on this side rationale.
\begin{equation}
    r^* = \argmax_{r} p(r \vert \mathcal{D},\overline{r}, f^*).
    \label{eq:response}
\end{equation}  

As a comparison, we also implemented an approach called \textsc{Syndicom-Multistep}. This approach breaks the inclusion of NLHF into two steps: (1) train a feedback model on NLHF that {\em predicts} the feedback critical of response $\overline{r}$
\begin{equation}
    \hat{f} = \argmax_{f} p(f \vert \mathcal{D},\overline{r}).
    \label{eq:multi1}
\end{equation}
and (2) train a second model to produce an improved dialogue response from the invalid response, given the {\it predicted} feedback 
\begin{equation}
    r^* = \argmax_{r} p(r \vert \mathcal{D},\overline{r}, \hat{f}).
    \label{eq:multi2}
\end{equation}  

\begin{table*}[]
    \centering
    \begin{tabular}{l c c c}
        \toprule
        \textbf{Description} & \textbf{Train} & \textbf{Val} & \textbf{Test} \\
        \midrule
        \# Samples & 16221 & 1709 & 1787 \\
        \# Turns per template & 5.21$\pm$1.42 & 5.26$\pm$1.42 & 5.23$\pm$1.42 \\
        \# Turns per dialogue & 5.18$\pm$1.36 & 5.21$\pm$1.36 & 5.18$\pm$1.32 \\
        \bottomrule
    \end{tabular}
    \caption{Statistics of our \textsc{Syndicom} dataset. \# Dialogue turns includes the valid response ($\pm$ indicates 1 std deviation.). The splits were inherited from \textsc{Atomic}, the source of the templates.}
    \label{tab:stats}
\end{table*}

\begin{table}[]
    \centering
    \begin{tabular}{l|c}
        \toprule
        \textbf{Hyperparameter} & \textbf{Value} \\
        \midrule
        Temperature & 0.7 \\
        Max tokens & 50 \\
        Top p & 1.0 \\
        Frequency penalty & 0 \\
        Presence penalty & 0 \\
        \bottomrule
    \end{tabular}
    \caption{Hyperparameters used for GPT-3.5. The same parameters were used for training and inference.}
    \label{tab:gpt}
\end{table}
Both models used in this work are based on OpenAI's GPT-3.5, specifically \lstinline{text-davinci-003}. The models were fine-tuned through the OpenAI API for GPT based models. The hyperparameters used are listed in Table \ref{tab:gpt}.

\section{Experiments}

In this section, we provide a detailed description of the experiments conducted to evaluate our proposed method, \textsc{Syndicom}. The experiments aim to compare the direct prediction of the improved response in Equation \ref{eq:direct} (\textsc{Syndicom-Direct}) with the response prediction when conditioned on natural language human feedback (NLHF) that explains why the initial response is invalid (\textsc{Syndicom-NLHF}). Additionally, we explore a multistep implementation of NLHF (\textsc{Syndicom-Multistep}). We compare the performance of our method against a ChatGPT baseline (\lstinline{gpt-3.5-turbo}) using various text generation metrics, such as ROUGE, BLEU, SacreBLEU, BERTScore, and METEOR.

\subsection{\textsc{Syndicom-Direct}}
Our first experiment focused on the direct dialogue improvement task, where the objective is to enhance a dialogue response based solely on the context and an invalid response. No feedback, whether human or generated, was involved in this task. This optimization problem is described in Equation \ref{eq:direct}.

In order to prevent the model from simply learning to undo the error injection, we introduced noise by rephrasing the invalid dialogues using an independent ChatGPT instance. This rephrasing was only performed at inference time and not during training. The rephrasing prompt is available in Appendix \ref{sec:appendix}.

\subsection{\textsc{Syndicom-Multistep}}
\begin{table*}[h]
    \centering
    \begin{tabular}{lccc|ccc}
    \toprule
    & \multicolumn{3}{c|}{\textbf{ChatGPT}} & \multicolumn{3}{c}{\textbf{\textsc{Syndicom}}} \\
    \cmidrule(lr){2-4} \cmidrule(lr){5-7}
    \textbf{Metric} & \textbf{Max} & \textbf{Min} & \textbf{Avg} & \textbf{Max} & \textbf{Min} & \textbf{Avg} \\
    \midrule
    \textbf{ROUGE1} & 0.204 & 0.123 & 0.163 & 0.315 & 0.185 & 0.250 \\
    \textbf{ROUGE2} & 0.034 & 0.0078 & 0.0209 & 0.112 & 0.035 & 0.073 \\
    \textbf{ROUGEL} & 0.150 & 0.093 & 0.122 & 0.248 & 0.144 & 0.196 \\
    \textbf{BERTSCORE} & 0.863 & 0.853 & 0.858 & 0.883 & 0.866 & 0.874 \\
    \textbf{SacreBLEU} & 2.546 & 1.533 & 2.039 & 6.697 & 2.907 & 4.802 \\
    \textbf{BLEU} & 0.004 & 0.0001 & 0.0021 & 0.030 & 0.0041 & 0.0171 \\
    \textbf{METEOR} & 0.197 & 0.129 & 0.163 & 0.279 & 0.158 & 0.219 \\
    \bottomrule
    \end{tabular}
    \caption{Performance in Feedback Generation performance of our method vs. baseline. \textsc{Syndicom} outperforms the baseline on all metrics. Each dialogue was accompanied by two feedback responses, and scores were computed for both independently. We show the max/min/avg over the two for each score and model.}
    \label{tab:feedback}
\end{table*}

Next, we explored the \textsc{Syndicom-Multistep} approach. As shown in Equations \ref{eq:multi1} and \ref{eq:multi2}, we first predicted feedback using the feedback model and then improved the dialogue response using the response improvement model. For the feedback predictor, we trained a GPT-based model to generate feedback given a dialogue context and an invalid response, as shown in Equation \ref{eq:multi1}, using the typical causal language modeling objective. We evaluated the feedback generation model portion of \textsc{Syndicom-Multistep} separately and compared it to ChatGPT. The prompt used for the baseline can be found in Appendix \ref{sec:appendix}. Table \ref{tab:feedback} presents the results, demonstrating that our method outperformed the baseline on all metrics.

Subsequently, we utilized the predicted feedback along with the dialogue context and invalid response to produce an improved dialogue response, as shown in Equation \ref{eq:multi2}. Similar to the \textsc{Syndicom-Direct} experiments, we applied rephrasing to the invalid responses at inference time. The baseline model was explicitly instructed to first generate feedback for the invalid response and then use that feedback to guide its response improvement. Table \ref{tab:rigf} displays the results.

\begin{table*}[h]
    \centering
    \begin{tabular}{lcc|cc|cc}
    \toprule
     & \multicolumn{2}{c|}{\textbf{ChatGPT}} & \multicolumn{3}{c}{\textbf{\textsc{Syndicom}}}\\
    \textbf{Metric} &Direct & NLHF & Direct & Multistep & NLHF \\
    \midrule
    \textbf{ROUGE1}    & 0.132 & 0.231 & 0.386  & \textbf{0.388} & \textit{0.474}  \\
    \textbf{ROUGE2}    & 0.029 & 0.081 & \textbf{0.174}  & 0.172 & \textit{0.246}  \\
    \textbf{ROUGEL}    & 0.112 & 0.201 & \textbf{0.324}  & 0.322 & \textit{0.396}  \\
    \textbf{BLEU}      & 0.008 & 0.031 & 0.117  & \textbf{0.125} & \textit{0.168}  \\
    \textbf{METEOR}    & 0.209 & 0.290 & \textbf{0.390}  & 0.387 & \textit{0.445}  \\
    \textbf{SacreBLEU} & 0.885 & 3.107 & 11.716 & \textbf{12.547}& \textit{16.831} \\
    \textbf{BERTScore} & 0.859 & 0.880 & 0.909  & \textbf{0.910} & \textit{0.919}  \\
    \bottomrule
    \end{tabular}
    \caption{Response Improvement comparing ChatGPT with our new \textsc{Syndicom} methods. ChatGPT-Direct is fine-tuned to produce a valid response given only the invalid response, with no intermediate steps or feedback. ChatGPT-NLHF is additionally conditioned on natural language human feedback (NLHF). \textsc{Syndicom-Direct} is the model that optimizes Equation \ref{eq:direct}, \textsc{Syndicom-Multistep} optimizes Equation \ref{eq:multi2}, and \textsc{Syndicom-NLHF} conditions on the same NLHF as used by the ChatGPT models. Bold text illustrates the highest score between all methods that are not give NLHF, and italics indicate the highest scores among NLHF tasks. \textsc{Syndicom} outperforms the baseline on all metrics for both tasks.}
    \label{tab:rigf}
\end{table*}

\subsection{\textsc{Syndicom-NLHF}}

The next experiment focused on enhancing dialogue responses using human feedback (Equation \ref{eq:response}). Given a dialogue context, an invalid response, and human feedback, the goal was to generate an improved (valid) dialogue response. For this experiment, we utilized the raw human-written feedback from \textsc{Syndicom} and trained a separate GPT improvement model to generate valid responses. As before, we applied inference-time rephrasing to the invalid responses. Results are presented in Table \ref{tab:rigf} under \textsc{Syndicom-NLHF}. This version of our method outperformed the others on all metrics.

\subsection{Human Evaluation}

In addition to our automated metric evaluations, we conducted a human evaluation to assess the effectiveness of response improvements through generated feedback. This evaluation process mirrored the dialogue enhancement steps employed in the experiment described in Section \ref{sec:dia_improvement}.

It is important to note that task assignments for crowdworkers require explicit and precise definitions, which often pose challenges in evaluating the commonsense aspect through human intervention. Existing human evaluations primarily focus on assessing the accuracy of information or determining the most preferred output from a set of alternatives.

With the emergence of advanced language models like ChatGPT, human evaluation has become increasingly complex. This complexity arises from the remarkably high-quality and naturally articulated outputs generated by state-of-the-art models such as ChatGPT.

In our study, we instructed crowdworkers that an AI system was attempting to emulate human conversation and generate dialogue responses that align with commonsense understanding and fit the given context. The workers were presented with two distinct responses: a standard ChatGPT response and our \textsc{Syndicom} response. Their task was to select the response that appeared more human-like and natural. The order of the responses chosen was randomized.

Despite the impressive contextual relevance exhibited by ChatGPT responses, our method generated the more favored response \textbf{56.5\%} of the time, compared to ChatGPT's 43.5\% preference rate. For further details on the interface provided to the crowdworkers, please refer to Appendix \ref{sec:appendix}.

\section{Discussion}

In the Discussion section, we analyze the performance of our proposed \textsc{Syndicom} method in conversational AI compared to the baseline model ChatGPT. The results are summarized in Tables \ref{tab:feedback} and \ref{tab:rigf}, where we observe that \textsc{Syndicom} outperforms ChatGPT on all automatic metrics for the feedback and dialogue response improvement tasks.

Specifically, Table \ref{tab:rigf} provides a comparison between our direct and multi-step approaches to the response improvement problem. Our multi-step method outperforms the direct method on various metrics such as ROUGE-1, BLEU, SacreBLEU, and BERTScore, despite the simplicity of the error typology used in the error injection during these experiments. This indicates that the multi-step approach has the potential to achieve even better performance when faced with more diverse error typologies, which we leave as an avenue for future research.

One contributing factor to the superior performance of the multi-step method is the additional information encoded in the feedback model. The feedback model is trained on human feedback, providing it with more contextual information compared to the direct model, which is solely trained on valid and invalid responses. Even in cases where the direct model achieves slightly higher scores in certain metrics, the differences are negligible. Notably, BERTScore, which represents the most comprehensive model-based metric utilized in our evaluation, further supports the argument in favor of the multi-step approach with feedback generation.

When examining the NLHF columns in Table \ref{tab:rigf}, we observe that \textsc{Syndicom} demonstrates significant improvement over ChatGPT for the response improvement task when provided with human feedback for the invalid response. This scenario aligns with use cases where feedback can be collected for a dialogue system and subsequently used to fine-tune and enhance the dialogue model. These findings underscore the value of the \textsc{Syndicom} method in continuous learning scenarios, particularly those where feedback from end users is actively being collected.

Overall, \textsc{Syndicom} exhibits strong performance compared to the state-of-the-art large language model ChatGPT, despite both models being based on the same underlying architecture (GPT-3.5). It is worth noting that ChatGPT underwent substantial reinforcement learning through human feedback during its refinement process, making the success of \textsc{Syndicom} even more noteworthy.
\section{Conclusion}

In this paper, we introduced \textsc{Syndicom}, a novel method for enhancing commonsense reasoning in dialogue response generation. By integrating a commonsense dialogue synthesis approach with targeted error injection, we tackled the challenge of incorporating commonsense knowledge into conversational AI systems. Our method comprised two key components: (1) a dataset consisting of valid and invalid responses to dialogue contexts, along with natural language feedback (NLF) for the invalid responses, and (2) a two-step procedure involving training a model to predict NLF for invalid responses, followed by training a response generation model conditioned on the predicted NLF, the invalid response, and the dialogue.

A notable advantage of \textsc{Syndicom} is its scalability and independence from reinforcement learning techniques, which are commonly employed in previous methods utilizing human feedback. Through comprehensive empirical evaluations across three tasks, we demonstrated the effectiveness of our approach using a diverse range of metrics. Notably, \textsc{Syndicom} outperformed ChatGPT on all metrics for both the dialogue improvement tasks, with and without human feedback.

To facilitate further research and practical adoption, we plan to release the code implementation of \textsc{Syndicom} as well as the complete dataset utilized in this work. By making these resources openly accessible, we aim to encourage collaboration and promote advancements in commonsense reasoning for dialogue systems.

\section*{Limitations and Future Work}
There are a few areas of limitation in this work. First, all the dialogues generated were based on templates synthesized from ATOMIC triplets. The domain is thus limited to the material contained in ATOMIC. Second, the procedural generation technique, while scaleable, inevitably introduces structure within the data that can be exploited by statistical models (including deep neural nets and language models). This is why the feedback generation task is particularly crucial, because the explanations are human-written and thus avoid such a limitation. 

Our experiments demonstrate our method of improving baseline dialogue responses that have been corrupted with error injection. This has the advantage of scale and targeting specific error modes that may be observed with LLMs, but the invalid responses in \textsc{Syndicom} do not themselves represent errors actually made by LLMs. A larger scale study could involve a data collection of errors and mistakes made by an LLM to demonstrate our method in improving baseline dialogue responses, but this approach would not lend itself to scale as any particular type of error made by state-of-the-art LLMs will likely be very rare. A more scaleable approach might be to develop a more comprehensive error typology and injection scheme, which we leave to future work. 

In future work, a more comprehensive error topology could be explored, along with a more substantial human evaluation, to explore the generalizability of the proposed method. This work focused on commonsense errors, but other errors that are observed in large language models could be explored in further analysis like mathematical reasoning, humor and sarcasm, etc. 



\bibliography{anthology,custom}

\begin{thebibliography}{29}
\expandafter\ifx\csname natexlab\endcsname\relax\def\natexlab#1{#1}\fi

\bibitem[{Bai et~al.(2022{\natexlab{a}})Bai, Jones, Ndousse, Askell, Chen,
  DasSarma, Drain, Fort, Ganguli, Henighan et~al.}]{bai2022training}
Yuntao Bai, Andy Jones, Kamal Ndousse, Amanda Askell, Anna Chen, Nova DasSarma,
  Dawn Drain, Stanislav Fort, Deep Ganguli, Tom Henighan, et~al.
  2022{\natexlab{a}}.
\newblock Training a helpful and harmless assistant with reinforcement learning
  from human feedback.
\newblock \emph{arXiv preprint arXiv:2204.05862}.

\bibitem[{Bai et~al.(2022{\natexlab{b}})Bai, Kadavath, Kundu, Askell, Kernion,
  Jones, Chen, Goldie, Mirhoseini, McKinnon et~al.}]{bai2022constitutional}
Yuntao Bai, Saurav Kadavath, Sandipan Kundu, Amanda Askell, Jackson Kernion,
  Andy Jones, Anna Chen, Anna Goldie, Azalia Mirhoseini, Cameron McKinnon,
  et~al. 2022{\natexlab{b}}.
\newblock Constitutional ai: Harmlessness from ai feedback.
\newblock \emph{arXiv preprint arXiv:2212.08073}.

\bibitem[{Brown et~al.(2020)Brown, Mann, Ryder, Subbiah, Kaplan, Dhariwal,
  Neelakantan, Shyam, Sastry, Askell et~al.}]{brown2020language}
Tom Brown, Benjamin Mann, Nick Ryder, Melanie Subbiah, Jared~D Kaplan, Prafulla
  Dhariwal, Arvind Neelakantan, Pranav Shyam, Girish Sastry, Amanda Askell,
  et~al. 2020.
\newblock Language models are few-shot learners.
\newblock \emph{Advances in neural information processing systems},
  33:1877--1901.

\bibitem[{Cui et~al.(2020)Cui, Wu, Liu, Zhang, and Zhou}]{cui2020mutual}
Leyang Cui, Yu~Wu, Shujie Liu, Yue Zhang, and Ming Zhou. 2020.
\newblock Mutual: A dataset for multi-turn dialogue reasoning.
\newblock \emph{arXiv preprint arXiv:2004.04494}.

\bibitem[{Ghosal et~al.(2022)Ghosal, Shen, Majumder, Mihalcea, and
  Poria}]{ghosal-etal-2022-cicero}
Deepanway Ghosal, Siqi Shen, Navonil Majumder, Rada Mihalcea, and Soujanya
  Poria. 2022.
\newblock \href {https://doi.org/10.18653/v1/2022.acl-long.344} {{CICERO}: A
  dataset for contextualized commonsense inference in dialogues}.
\newblock In \emph{Proceedings of the 60th Annual Meeting of the Association
  for Computational Linguistics (Volume 1: Long Papers)}, pages 5010--5028,
  Dublin, Ireland. Association for Computational Linguistics.

\bibitem[{Hwang et~al.(2021)Hwang, Bhagavatula, Le~Bras, Da, Sakaguchi,
  Bosselut, and Choi}]{hwang2021comet}
Jena~D Hwang, Chandra Bhagavatula, Ronan Le~Bras, Jeff Da, Keisuke Sakaguchi,
  Antoine Bosselut, and Yejin Choi. 2021.
\newblock (comet-) atomic 2020: On symbolic and neural commonsense knowledge
  graphs.
\newblock In \emph{Proceedings of the AAAI Conference on Artificial
  Intelligence}, volume~35, pages 6384--6392.

\bibitem[{Kim et~al.(2022)Kim, Yu, Jiang, Lu, Khashabi, Kim, Choi, and
  Sap}]{kim2022prosocialdialog}
Hyunwoo Kim, Youngjae Yu, Liwei Jiang, Ximing Lu, Daniel Khashabi, Gunhee Kim,
  Yejin Choi, and Maarten Sap. 2022.
\newblock Prosocialdialog: A prosocial backbone for conversational agents.
\newblock \emph{arXiv preprint arXiv:2205.12688}.

\bibitem[{Li et~al.(2017)Li, Su, Shen, Li, Cao, and Niu}]{li2017dailydialog}
Yanran Li, Hui Su, Xiaoyu Shen, Wenjie Li, Ziqiang Cao, and Shuzi Niu. 2017.
\newblock Dailydialog: A manually labelled multi-turn dialogue dataset.
\newblock \emph{arXiv preprint arXiv:1710.03957}.

\bibitem[{Liu et~al.(2017)Liu, Tur, Hakkani-Tur, Shah, and Heck}]{liu2017end}
Bing Liu, Gokhan Tur, Dilek Hakkani-Tur, Pararth Shah, and Larry Heck. 2017.
\newblock End-to-end optimization of task-oriented dialogue model with deep
  reinforcement learning.
\newblock \emph{Conversational AI Workshop, Neural Information Processing
  Systems (NeurIPS)}.

\bibitem[{Liu et~al.(2018)Liu, T{\"u}r, Hakkani-T{\"u}r, Shah, and
  Heck}]{liu-etal-2018-dialogue}
Bing Liu, Gokhan T{\"u}r, Dilek Hakkani-T{\"u}r, Pararth Shah, and Larry Heck.
  2018.
\newblock \href {https://doi.org/10.18653/v1/N18-1187} {Dialogue learning with
  human teaching and feedback in end-to-end trainable task-oriented dialogue
  systems}.
\newblock In \emph{Proceedings of the 2018 Conference of the North {A}merican
  Chapter of the Association for Computational Linguistics: Human Language
  Technologies, Volume 1 (Long Papers)}, pages 2060--2069, New Orleans,
  Louisiana. Association for Computational Linguistics.

\bibitem[{Lowe et~al.(2015)Lowe, Pow, Serban, and Pineau}]{lowe2015ubuntu}
Ryan Lowe, Nissan Pow, Iulian Serban, and Joelle Pineau. 2015.
\newblock The ubuntu dialogue corpus: A large dataset for research in
  unstructured multi-turn dialogue systems.
\newblock \emph{arXiv preprint arXiv:1506.08909}.

\bibitem[{Moon et~al.(2019)Moon, Shah, Kumar, and Subba}]{moon2019opendialkg}
Seungwhan Moon, Pararth Shah, Anuj Kumar, and Rajen Subba. 2019.
\newblock Opendialkg: Explainable conversational reasoning with attention-based
  walks over knowledge graphs.
\newblock In \emph{Proceedings of the 57th Annual Meeting of the Association
  for Computational Linguistics}, pages 845--854.

\bibitem[{Nayak et~al.(2017)Nayak, Hakkani-T{\"u}r, Walker, and
  Heck}]{nayak2017plan}
Neha Nayak, Dilek Hakkani-T{\"u}r, Marilyn~A Walker, and Larry~P Heck. 2017.
\newblock To plan or not to plan? discourse planning in slot-value informed
  sequence to sequence models for language generation.
\newblock In \emph{INTERSPEECH}, pages 3339--3343.

\bibitem[{Ouyang et~al.(2022)Ouyang, Wu, Jiang, Almeida, Wainwright, Mishkin,
  Zhang, Agarwal, Slama, Ray et~al.}]{ouyang2022training}
Long Ouyang, Jeffrey Wu, Xu~Jiang, Diogo Almeida, Carroll Wainwright, Pamela
  Mishkin, Chong Zhang, Sandhini Agarwal, Katarina Slama, Alex Ray, et~al.
  2022.
\newblock Training language models to follow instructions with human feedback.
\newblock \emph{Advances in Neural Information Processing Systems},
  35:27730--27744.

\bibitem[{Qin et~al.(2021)Qin, Gupta, Upadhyay, He, Choi, and
  Faruqui}]{qin2021timedial}
Lianhui Qin, Aditya Gupta, Shyam Upadhyay, Luheng He, Yejin Choi, and Manaal
  Faruqui. 2021.
\newblock Timedial: Temporal commonsense reasoning in dialog.
\newblock \emph{arXiv preprint arXiv:2106.04571}.

\bibitem[{Richardson and Heck(2023)}]{richardson2023commonsense}
Christopher Richardson and Larry Heck. 2023.
\newblock Commonsense reasoning for conversational ai: A survey of the state of
  the art.
\newblock \emph{arXiv preprint arXiv:2302.07926}.

\bibitem[{Sap et~al.(2019)Sap, Le~Bras, Allaway, Bhagavatula, Lourie, Rashkin,
  Roof, Smith, and Choi}]{sap2019atomic}
Maarten Sap, Ronan Le~Bras, Emily Allaway, Chandra Bhagavatula, Nicholas
  Lourie, Hannah Rashkin, Brendan Roof, Noah~A Smith, and Yejin Choi. 2019.
\newblock Atomic: An atlas of machine commonsense for if-then reasoning.
\newblock In \emph{Proceedings of the AAAI conference on artificial
  intelligence}, volume~33, pages 3027--3035.

\bibitem[{Saunders et~al.(2022)Saunders, Yeh, Wu, Bills, Ouyang, Ward, and
  Leike}]{saunders2022self}
William Saunders, Catherine Yeh, Jeff Wu, Steven Bills, Long Ouyang, Jonathan
  Ward, and Jan Leike. 2022.
\newblock Self-critiquing models for assisting human evaluators.
\newblock \emph{arXiv preprint arXiv:2206.05802}.

\bibitem[{Scheurer et~al.(2022)Scheurer, Campos, Chan, Chen, Cho, and
  Perez}]{scheurer2022training}
J{\'e}r{\'e}my Scheurer, Jon~Ander Campos, Jun~Shern Chan, Angelica Chen,
  Kyunghyun Cho, and Ethan Perez. 2022.
\newblock Training language models with natural language feedback.
\newblock \emph{arXiv preprint arXiv:2204.14146}.

\bibitem[{Shah et~al.(2016)Shah, Hakkani-T{\"u}r, T{\"u}r, and
  Heck}]{shah2016interactive}
Pararth Shah, Dilek Hakkani-T{\"u}r, T{\"u}r, and Larry Heck. 2016.
\newblock Interactive reinforcement learning for task-oriented dialogue
  management.
\newblock \emph{Workshop on Deep Learning for Action and Interaction, Neural
  Information Processing Systems (NIPS)}.

\bibitem[{Speer et~al.(2017)Speer, Chin, and Havasi}]{speer2017conceptnet}
Robyn Speer, Joshua Chin, and Catherine Havasi. 2017.
\newblock Conceptnet 5.5: An open multilingual graph of general knowledge.
\newblock In \emph{Proceedings of the AAAI conference on artificial
  intelligence}, volume~31.

\bibitem[{Stiennon et~al.(2020)Stiennon, Ouyang, Wu, Ziegler, Lowe, Voss,
  Radford, Amodei, and Christiano}]{stiennon2020learning}
Nisan Stiennon, Long Ouyang, Jeffrey Wu, Daniel Ziegler, Ryan Lowe, Chelsea
  Voss, Alec Radford, Dario Amodei, and Paul~F Christiano. 2020.
\newblock Learning to summarize with human feedback.
\newblock \emph{Advances in Neural Information Processing Systems},
  33:3008--3021.

\bibitem[{Sun et~al.(2022)Sun, Zhang, Mi, Wang, Liu, Cui, Wang, Liu, and
  Huang}]{sun2022moraldial}
Hao Sun, Zhexin Zhang, Fei Mi, Yasheng Wang, Wei Liu, Jianwei Cui, Bin Wang,
  Qun Liu, and Minlie Huang. 2022.
\newblock Moraldial: A framework to train and evaluate moral dialogue systems
  via constructing moral discussions.
\newblock \emph{arXiv preprint arXiv:2212.10720}.

\bibitem[{Sun et~al.(2019)Sun, Yu, Chen, Yu, Choi, and Cardie}]{sun2019dream}
Kai Sun, Dian Yu, Jianshu Chen, Dong Yu, Yejin Choi, and Claire Cardie. 2019.
\newblock Dream: A challenge data set and models for dialogue-based reading
  comprehension.
\newblock \emph{Transactions of the Association for Computational Linguistics},
  7:217--231.

\bibitem[{Welleck et~al.(2022)Welleck, Lu, West, Brahman, Shen, Khashabi, and
  Choi}]{welleck2022generating}
Sean Welleck, Ximing Lu, Peter West, Faeze Brahman, Tianxiao Shen, Daniel
  Khashabi, and Yejin Choi. 2022.
\newblock Generating sequences by learning to self-correct.
\newblock \emph{arXiv preprint arXiv:2211.00053}.

\bibitem[{Weston et~al.(2018)Weston, Dinan, and Miller}]{weston2018retrieve}
Jason Weston, Emily Dinan, and Alexander~H Miller. 2018.
\newblock Retrieve and refine: Improved sequence generation models for
  dialogue.
\newblock \emph{arXiv preprint arXiv:1808.04776}.

\bibitem[{Zhou et~al.(2021{\natexlab{a}})Zhou, Gopalakrishnan, Hedayatnia, Kim,
  Pujara, Ren, Liu, and Hakkani-Tur}]{zhou2021commonsense}
Pei Zhou, Karthik Gopalakrishnan, Behnam Hedayatnia, Seokhwan Kim, Jay Pujara,
  Xiang Ren, Yang Liu, and Dilek Hakkani-Tur. 2021{\natexlab{a}}.
\newblock Commonsense-focused dialogues for response generation: An empirical
  study.
\newblock \emph{arXiv preprint arXiv:2109.06427}.

\bibitem[{Zhou et~al.(2021{\natexlab{b}})Zhou, Deshmukh, Greer, and
  Lee}]{zhou2021narle}
Ruijie Zhou, Soham Deshmukh, Jeremiah Greer, and Charles Lee.
  2021{\natexlab{b}}.
\newblock Narle: Natural language models using reinforcement learning with
  emotion feedback.
\newblock \emph{arXiv preprint arXiv:2110.02148}.

\bibitem[{Ziems et~al.(2022)Ziems, Yu, Wang, Halevy, and Yang}]{ziems2022moral}
Caleb Ziems, Jane~A Yu, Yi-Chia Wang, Alon Halevy, and Diyi Yang. 2022.
\newblock The moral integrity corpus: A benchmark for ethical dialogue systems.
\newblock \emph{arXiv preprint arXiv:2204.03021}.

\end{thebibliography}
\bibliographystyle{acl_natbib}

\appendix

\section{GPT-3 Prompts and Mechanical Turk interfaces}
\label{sec:appendix}

\begin{table*}[]
    \centering
    \begin{tabular}{p{1.5in}|p{4in}}
        \toprule
       \textbf{Task} & \textbf{Prompt} \\
       \midrule
       Direct & You will be given a dialogue context and a baseline     response. Your job is to improve that baseline response.       Always write the improved response last and prefix it with     'Improved Response:' \\ \midrule
       NLHF & You will be given a dialogue context and a baseline response. Your job is to improve that baseline response. Do so by first generating feedback for that response, as if it was written by an AI and you are critiquing it, and then produce the improved response. Always write the improved response last and prefix it with 'Improved Response:' \\ \midrule
       Feedback Generation & You are shown a synthetic dialogue written by an AI. The dialogue is intended to sound like a natural text message conversation between two people. The AI is imperfect and makes mistakes. You are asked to provide feedback to the AI to improve its dialogue generation. You are given a few dialogue turns, followed by a Baseline Response. Please give 1-2 sentences of feedback for the baseline response, and please be specific! \\
        \bottomrule
    \end{tabular}
    \caption{Prompts used for ChatGPT baselines}
    \label{tab:prompts}
\end{table*}

\begin{figure*}
    \centering
    \includegraphics[width=\textwidth]{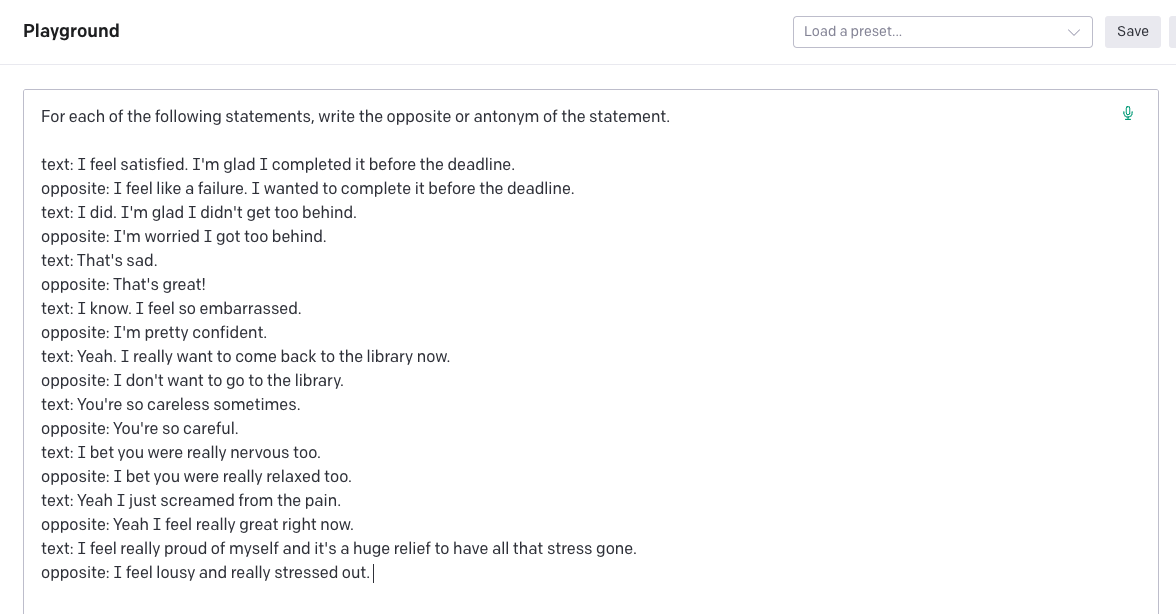}
    \caption{GPT-3 Prompt used for creating invalid dialogue responses from valid responses.}
    \label{fig:prompt_negations}
\end{figure*}

\begin{figure*}
    \centering
    \includegraphics[width=\textwidth]{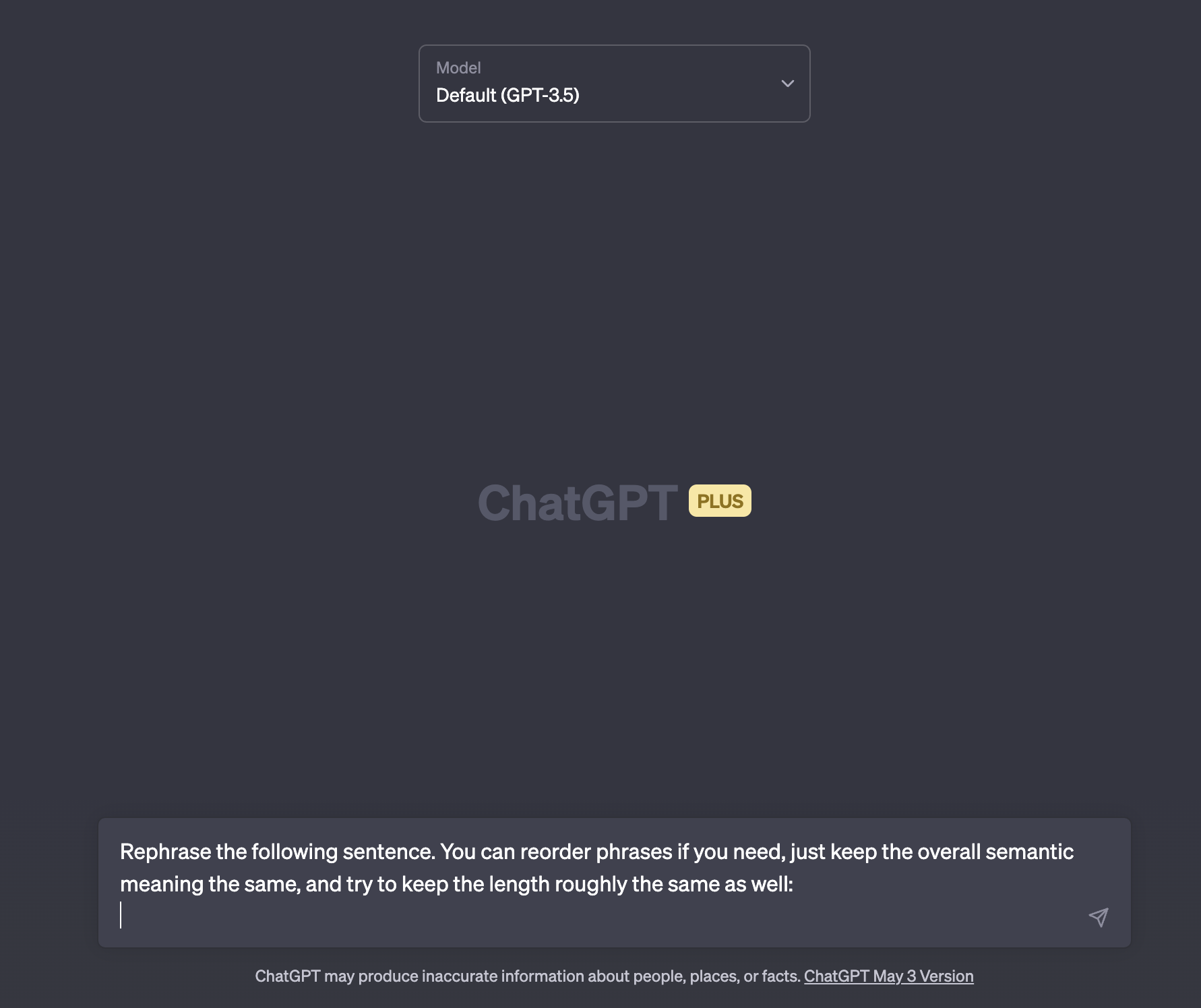}
    \caption{ChatGPT prompt used for rephrasing invalid dialogue responses.}
    \label{fig:prompt_rephrase}
\end{figure*}

\begin{figure*}
    \centering
    \includegraphics[width=\textwidth]{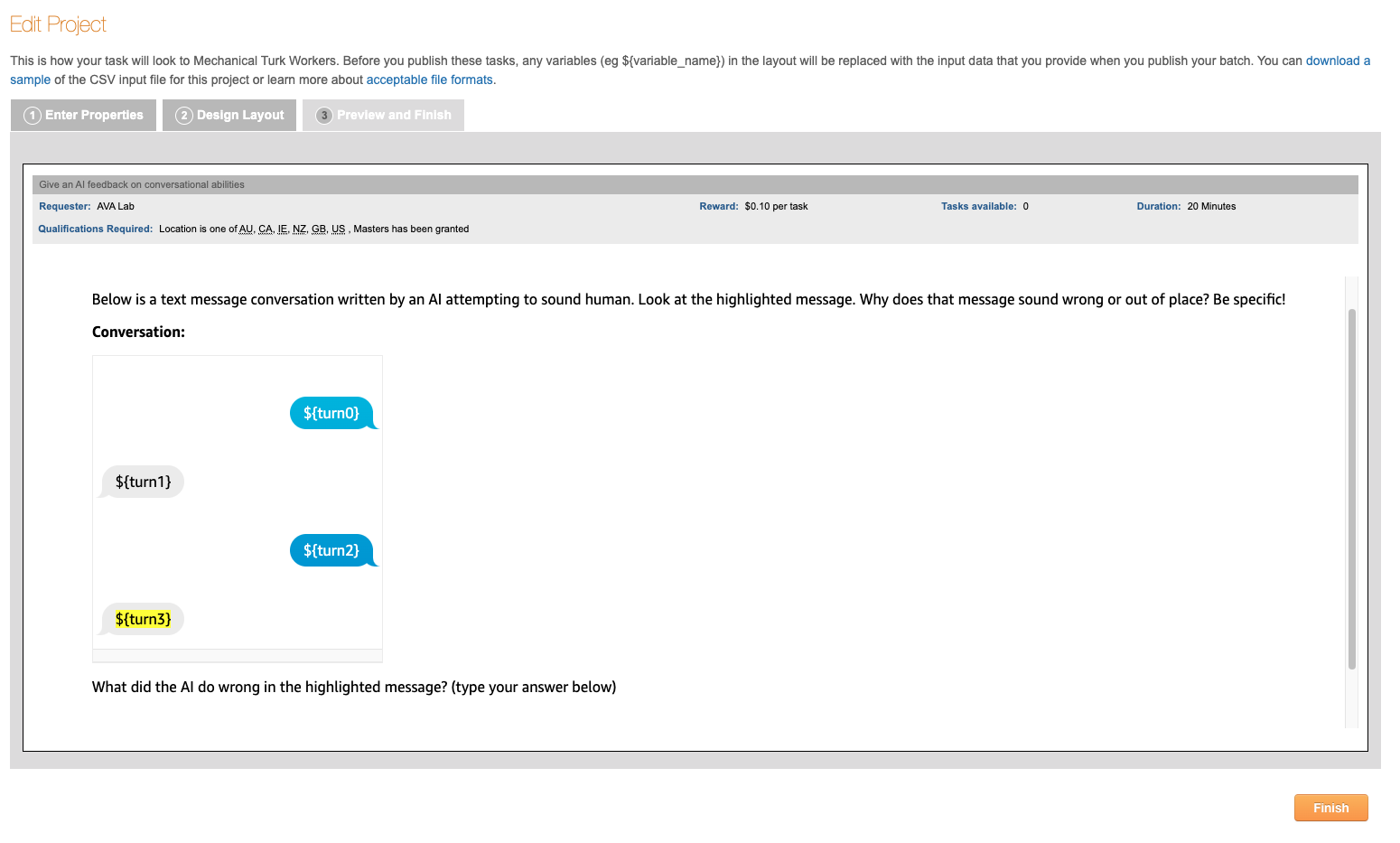}
    \caption{Mechanical Turk interface used for acquiring feedback for dialogue responses. Each dialogue was given feedback by two independent crowdworkers.}
    \label{fig:app_feedback}
\end{figure*}

\begin{figure*}
    \centering
    \includegraphics[width=\textwidth]{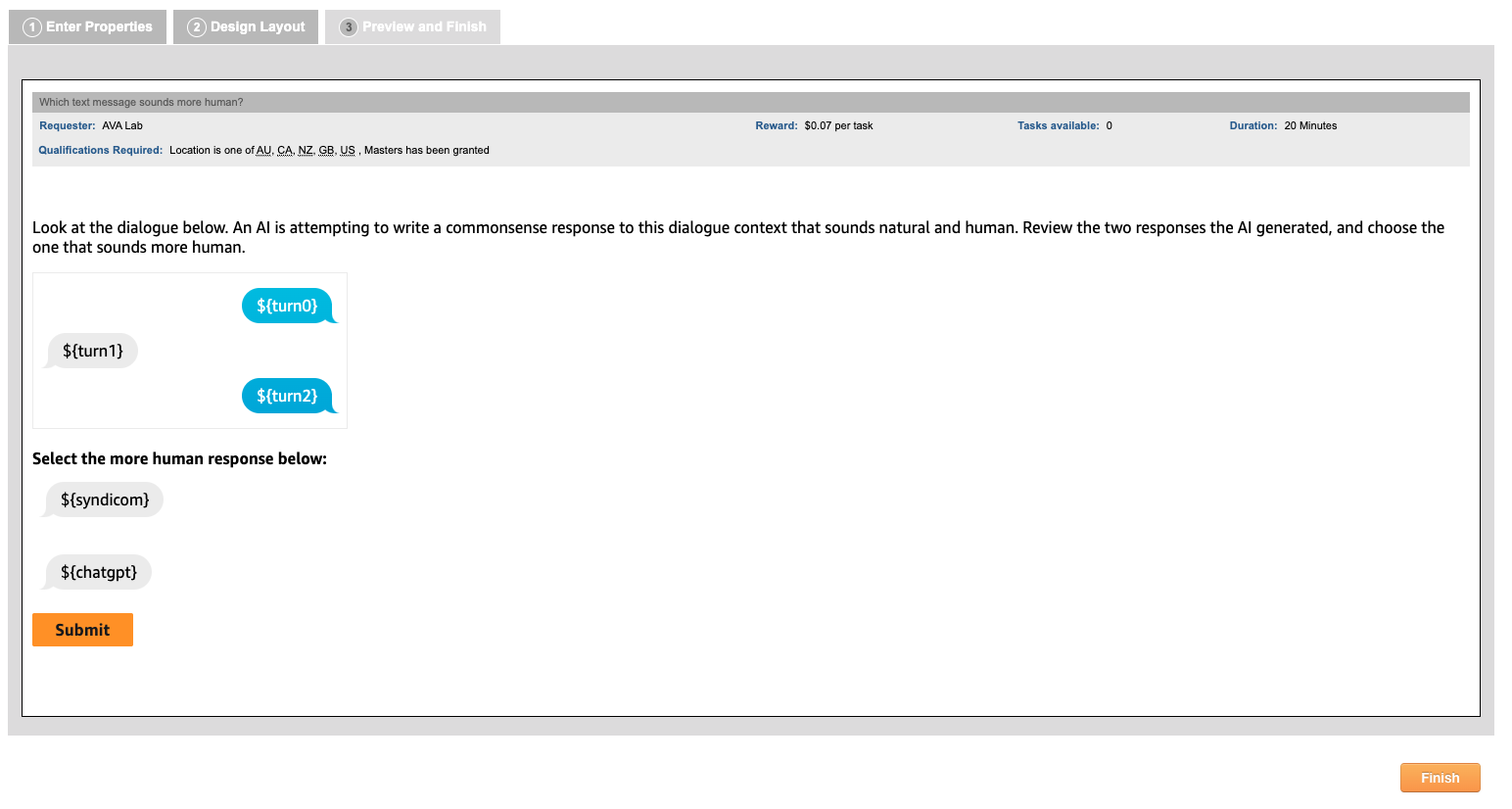}
    \caption{Mechanical Turk interface used for human evaluation. Each dialogue response pair was evaluated by two workers independently. Templates are shown instead of examples in order to fit the page.}
    \label{fig:app_evaluation}
\end{figure*}

\end{document}